\documentclass{article}
\usepackage{iclr2027_conference,times}
\iclrfinalcopy

\usepackage{amsmath,amsfonts,bm}

\def\eqref#1{equation~\ref{#1}}

\def\1{\bm{1}}

\DeclareMathAlphabet{\mathsfit}{\encodingdefault}{\sfdefault}{m}{sl}
\SetMathAlphabet{\mathsfit}{bold}{\encodingdefault}{\sfdefault}{bx}{n}

\usepackage{hyperref}
\usepackage{url}
\usepackage{amsmath,amssymb}
\usepackage{booktabs}
\usepackage{graphicx}
\usepackage{algorithm}
\usepackage{algorithmic}

\newcommand{\relucu}[1]{(#1)_+^3}

\title{InKAN: B-Spline KANs via Truncated Power Form}

\author{Naveen Mysore \\
\texttt{nmysore.work@gmail.com}}

\begin{document}
\maketitle

\begin{abstract}

Kolmogorov-Arnold Networks (KANs) place learnable B-spline activations on
network edges rather than fixed activations on nodes. The standard Cox-de
Boor recursion evaluates these activations through $k$ sequential passes
for degree-$k$ splines, consuming over 90\% of forward-pass time.

InKAN replaces this recursion with the truncated power form, a classical
result from approximation theory that expresses each uniform cubic
B-spline as five $(x)_+^3$ terms at shifted knot positions. The
resulting expression computes exact B-spline basis values: the same
mathematical function as the Cox-de Boor recursion, evaluated without
sequential passes. This paper documents three contributions: (1) an
implementation structured for \texttt{torch.compile} fusion, eliminating
recursion, span lookup, and scatter-gather operations; (2) a
bounded-coordinate evaluation that clamps the normalized input to
$[0, k{+}1]$, preventing the growth of cancellation error at large
off-support coordinates; and (3) an open-source package
(\texttt{pip install inkan}). In the tested configurations, InKAN has
2.8--3.5$\times$ lower forward-pass latency than the Cox-de Boor
recursion. Partition-of-unity errors remain below $10^{-5}$ for grid
sizes up to 200.

\end{abstract}

\section{Introduction}
\label{sec:intro}

The Kolmogorov-Arnold representation theorem~\citep{kolmogorov1957representation}
guarantees that any multivariate continuous function decomposes into
univariate functions and addition. KANs~\citep{liu2024kan} operationalize
this result by placing a learnable activation on every edge. Each activation is a
weighted sum of B-spline basis functions, which can be inspected,
plotted, and in favorable cases symbolically recovered.

This interpretability carries a cost. The Cox-de Boor
recursion~\citep{cox1972numerical, deboor1971subroutine} evaluates
B-spline basis functions through $k$ sequential passes for degree $k$.
For cubic splines ($k{=}3$), three passes are needed, each depending on
the output of the previous. In the profiled configuration (Table~\ref{tab:cost_breakdown}), basis
computation accounts for 91\% of a KAN layer's forward-pass time.

The recursion itself was not always the standard. In the 1940s,
Schoenberg introduced splines as piecewise polynomials while smoothing
ballistics data at Aberdeen Proving
Ground~\citep{schoenberg1946contributions}. His representation used
truncated power functions $\max(0, x)^k$ as the foundational building
block, and Curry and Schoenberg formalized B-splines through this
basis~\citep{curry1947splines}. The truncated power form is
mathematically elegant: proofs of compact support, partition of unity,
and smoothness follow directly. But early implementations on 1960s
hardware suffered from numerical cancellation in the alternating sum of
large powers, and the truncated power basis matrix is poorly
conditioned~\citep{deboor2001practical, schumaker2007spline}. Cox and
de Boor independently developed the recursive
algorithm~\citep{cox1972numerical, deboor1971subroutine} as a
numerically stable alternative, and de Boor's
textbook~\citep{deboor2001practical} established the recursion as the
standard for the next five decades.

Prior work has pursued several strategies to reduce the cost of B-spline
evaluation in KANs. Restructuring the Cox-de Boor
recursion~\citep{blealtan2024} reduces constant factors but preserves
the sequential dependency. Matrix-form
evaluation~\citep{coffman2025matrixkan} eliminates the recursion but
introduces data-dependent gather operations. Gaussian radial basis
functions~\citep{li2024fastkan} replace B-splines entirely with a
single \texttt{exp()} call, gaining speed but sacrificing compact
support and automatic partition of unity (Gaussians are non-negative
and $C^\infty$, so smoothness is not the trade-off).
PowerMLP~\citep{qiu2025powermlp} connects B-splines to powers of
rectified linear units and develops a non-iterative alternative, but
replaces the spline parameterization rather than evaluating it directly.
\citet{southworth2026multilevel} established the algebraic equivalence
between spline KAN layers and power-ReLU networks through a
change-of-basis matrix. Their emphasis is multilevel training; InKAN
documents a particular implementation, stabilization procedure, and
benchmarks for the same algebraic identity.

InKAN bridges the gap between algebraic equivalence and practical
deployment. The contributions are:

\begin{enumerate}
\item \textbf{Compiler-fused evaluation.} For uniform cubic B-splines,
the truncated power form reduces to five $\max(0,\cdot)^3$ terms with
fixed coefficients. The elementwise expression is structured for
\texttt{torch.compile} fusion, eliminating recursion, span lookup,
and data-dependent memory access.

\item \textbf{Bounded-coordinate evaluation.} The raw truncated power
sum suffers from growing cancellation error when the normalized input $u$
lies far outside the basis support $[0, k{+}1]$. Clamping $u$ to this
interval before evaluation bounds all intermediate terms, preventing the
$O(\epsilon_{\mathrm{mach}} \cdot u^3)$ error growth, with no change to
the mathematically correct output (since the B-spline is exactly zero
outside its support). Small floating-point errors remain within the
support; their magnitude is quantified in Table~\ref{tab:stability}.

\item \textbf{Open-source package.} InKAN is available as an
open-source Python package (\texttt{pip install inkan}, MIT license)
providing PyTorch-compatible KAN layers for fixed, uniform cubic
spline grids.
\end{enumerate}

\section{From Recursion to Closed Form}
\label{sec:method}

The derivation proceeds in four steps. Each transforms an algorithm into
an expression, revealing structure that the algorithm conceals.

\subsection{Step 1: De Casteljau to Bernstein}

Let $P_0, P_1, \ldots, P_n \in \mathbb{R}^d$ be a sequence of
\emph{control points} and let $t \in [0, 1]$ be a scalar parameter. The
\emph{linear interpolation} (lerp) between two points is defined as
$\operatorname{lerp}(A, B, t) = (1 - t)\,A + t\,B$. De Casteljau's
algorithm~\citep{farin2002curves} evaluates a degree-$n$ B\'ezier
curve $Q(t)$ by applying $\operatorname{lerp}$ recursively: at each
level, adjacent points are interpolated pairwise, reducing the number of
points by one until a single value remains.

For the cubic case ($n{=}3$), three levels of interpolation over
$P_0, P_1, P_2, P_3$ produce the curve point $Q(t)$. Expanding and
collecting terms on each control point yields the Bernstein form:
\begin{equation}
\label{eq:bernstein}
Q(t) = \sum_{i=0}^{3} B_{i,3}(t)\, P_i, \quad
B_{i,3}(t) = \binom{3}{i}\, t^i\, (1{-}t)^{3-i}
\end{equation}
The coefficients $\binom{3}{i} = \{1, 3, 3, 1\}$ are row~3 of Pascal's
triangle. The partition of unity $\sum_{i=0}^{3} B_{i,3}(t) = 1$
follows from $((1{-}t) + t)^3 = 1$.

\subsection{Step 2: Matrix form}

Expanding each Bernstein polynomial into powers of $t$ separates the
B\'ezier curve into three factors:
\begin{equation}
\label{eq:bezier_matrix}
Q(t) = \underbrace{\begin{bmatrix} 1 & t & t^2 & t^3 \end{bmatrix}}_{\mathbf{T}(t)}
\underbrace{\begin{bmatrix}
  1 & 0 & 0 & 0 \\
 -3 & 3 & 0 & 0 \\
  3 &-6 & 3 & 0 \\
 -1 & 3 &-3 & 1
\end{bmatrix}}_{\mathbf{M}_{\mathrm{B\acute{e}zier}}}
\underbrace{\begin{bmatrix} P_0 \\ P_1 \\ P_2 \\ P_3 \end{bmatrix}}_{\mathbf{P}}
\end{equation}
The factorization $\mathbf{T}(t)\,\mathbf{M}\,\mathbf{P}$ is general.
Different spline types share the same structure but differ only in
$\mathbf{M}$. The input-dependent computation is always four values
$[1, t, t^2, t^3]$, trivially cheap.

\subsection{Step 3: B\'ezier segments to B-splines}

A single cubic B\'ezier has a fundamental limitation: \emph{global}
control. Moving any control point affects the entire curve. To model a
function over a wide range, multiple segments must be chained, and at
every joint three constraints must be enforced ($C^0$, $C^1$, $C^2$
continuity). The constraint cost grows linearly with the number of
segments.

B-splines solve this by construction. Given a non-decreasing \emph{knot
vector} $\mathbf{t} = (t_0, t_1, \ldots, t_m)$, the Cox-de Boor
recursion~\citep{deboor2001practical} defines basis functions
$N_{i,k}(x)$ of degree $k$. For the uniformly spaced, simple knots
used here, these are $C^{k-1}$ across interior knots:
\begin{align}
N_{i,0}(x) &= \begin{cases} 1 & \text{if } t_i \leq x < t_{i+1} \\ 0 & \text{otherwise} \end{cases} \label{eq:cdb0} \\[4pt]
N_{i,k}(x) &= \frac{x - t_i}{t_{i+k} - t_i}\, N_{i,k-1}(x)
             + \frac{t_{i+k+1} - x}{t_{i+k+1} - t_{i+1}}\, N_{i+1,k-1}(x) \label{eq:cdbk}
\end{align}
Each basis function has \emph{compact support}: $N_{i,k}(x)$ is nonzero
only over $k{+}1$ consecutive knot spans. For a uniform knot vector
with spacing $h$, the B-spline basis matrix is:
\begin{equation}
\label{eq:bspline_matrix}
\mathbf{M}_{\mathrm{B\text{-}spline}} = \frac{1}{6}\begin{bmatrix}
  1 & 4 & 1 & 0 \\
 -3 & 0 & 3 & 0 \\
  3 &-6 & 3 & 0 \\
 -1 & 3 &-3 & 1
\end{bmatrix}
\end{equation}

For $k{=}3$, three sequential passes are needed. Each depends on the
previous. Table~\ref{tab:cost_breakdown} shows these passes consume
91\% of a KAN layer's forward time.

\begin{table}[t]
\caption{Forward-pass cost breakdown for a KAN layer
(256$\times$784$\to$64, MPS GPU). Basis computation dominates.}
\label{tab:cost_breakdown}
\begin{center}
\begin{tabular}{lrr}
\toprule
Component & Time (ms) & Share \\
\midrule
B-spline basis (Cox-de Boor, 3 passes) & 1.44 & 91\% \\
Spline \texttt{einsum} (bases $\times$ weights) & 0.09 & 6\% \\
Base function (SiLU + \texttt{einsum}) & 0.05 & 3\% \\
\midrule
Total forward & 1.58 & 100\% \\
\bottomrule
\end{tabular}
\end{center}
\end{table}

\subsection{Step 4: Cox-de Boor to truncated power form}

For a \emph{uniform} knot vector with constant spacing
$h = t_{i+1} - t_i$, expanding the recursion yields a finite difference
of truncated power
functions~\citep{curry1947splines, schoenberg1946contributions}:
\begin{equation}
\label{eq:general_truncated}
N_{i,k}(x) = \frac{1}{k!\, h^k} \sum_{j=0}^{k+1} (-1)^j \binom{k{+}1}{j}\, \max(0,\; x - t_{i+j})^k
\end{equation}
For cubic splines ($k{=}3$), substituting $u = (x - g_i)/h$:
\begin{equation}
\label{eq:truncated}
N_i(u) = \frac{1}{6}\Big[\relucu{u} - 4\relucu{u{-}1}
  + 6\relucu{u{-}2} - 4\relucu{u{-}3} + \relucu{u{-}4}\Big]
\end{equation}
where $g_i$ is the start of the $i$-th support interval and
$(z)_+ = \max(0, z)$. The coefficients $\{1, -4, 6, -4, 1\}$ are the
fourth row of Pascal's triangle with alternating signs, divided by
$3! = 6$. This is the same coefficient stencil identified by
\citet{southworth2026multilevel} through their change-of-basis analysis.

De Casteljau is to Bernstein as Cox-de Boor is to the truncated power
form. In both cases, an algorithm built from nested interpolations
admits a closed-form expression built from binomial coefficients. The
algorithm is sequential. The expression is parallel and structured
for compiler fusion.

\subsection{Bounded-coordinate stabilization}
\label{sec:bounded}

In finite-precision arithmetic, the alternating sum of five terms of
size $\Theta(u^3)$ produces growing cancellation error when $u$ lies
far outside the support interval $[0, 4]$. The mathematically correct
value is exactly zero, but floating-point evaluation produces a residual
that grows as $O(\epsilon_{\mathrm{mach}} \cdot u^3)$. This growth was
among the concerns that motivated the Cox-de Boor
recursion~\citep{cox1972numerical, deboor1971subroutine,
deboor2001practical}.

Before evaluating Eq.~\ref{eq:truncated}, InKAN clamps the normalized
coordinate to the support:
\begin{equation}
\label{eq:bounded}
\bar{u} = \operatorname{clamp}(u,\; 0,\; k{+}1)
\end{equation}
This is algebraically correct ($N_i(u) = 0$ outside $[0, k{+}1]$) and
bounds every intermediate term: the largest cubed quantity is at most
$(k{+}1)^3 = 64$ for cubic splines. Table~\ref{tab:stability} shows
the effect across grid sizes.

\begin{table}[t]
\caption{Maximum out-of-support residual
$\max_{x,i\notin S_i}|\widehat{N}_i(x)|$ and partition-of-unity error
$\max_x|\sum_i \widehat{N}_i(x) - 1|$ for unclamped and
bounded-coordinate evaluation. 2,001 inputs in $[-1, 1]$, float32,
CPU, eager execution.}
\label{tab:stability}
\begin{center}
\begin{tabular}{lrrrr}
\toprule
& \multicolumn{2}{c}{Without clamp} & \multicolumn{2}{c}{With clamp} \\
\cmidrule(lr){2-3} \cmidrule(lr){4-5}
$G$ & Out-of-support & PU error & Out-of-support & PU error \\
\midrule
5 & $3.8 \times 10^{-5}$ & $5.3 \times 10^{-5}$ & 0 & $2.2 \times 10^{-6}$ \\
32 & $4.9 \times 10^{-3}$ & $1.8 \times 10^{-2}$ & 0 & $1.7 \times 10^{-6}$ \\
64 & $4.2 \times 10^{-2}$ & $2.5 \times 10^{-1}$ & 0 & $1.2 \times 10^{-6}$ \\
100 & $1.7 \times 10^{-1}$ & $8.8 \times 10^{-1}$ & 0 & $4.2 \times 10^{-6}$ \\
200 & --- & --- & 0 & $8.6 \times 10^{-6}$ \\
\bottomrule
\end{tabular}
\end{center}
\end{table}

\subsection{Integration with KAN layers}

A KAN layer computes:
\begin{equation}
\label{eq:kan_forward}
y_o = \sum_i \left[\sum_k c_{o,i,k}\, N_k(x_i) + w_{o,i}\, \sigma(x_i)\right]
\end{equation}
where $c_{o,i,k}$ are learnable spline coefficients, $w_{o,i}$ are
residual weights, and $\sigma$ is SiLU. Replacing the Cox-de Boor
evaluation of $N_k(x_i)$ with the bounded truncated power form
(Eq.~\ref{eq:truncated} with Eq.~\ref{eq:bounded}) changes only the
basis computation; the same mathematical mapping is preserved under
matched knots and effective coefficients.

\section{Experiments}
\label{sec:experiments}

InKAN is evaluated on three axes: \textbf{speed}, \textbf{classification},
and \textbf{regression}. All KAN variants use grid\_size=5,
spline\_order=3, architecture input$\to$64$\to$10 (classification) or
input$\to$8$\to$1 (regression), Adam with lr=$10^{-3}$, and batch size
256. Classification and regression use 10 random seeds (42--51) and
report mean $\pm$ standard deviation. The MLP baseline uses
Linear+SiLU+Linear with matched layer dimensions.

\subsection{Speed}

Table~\ref{tab:speed} reports forward-pass latency on NVIDIA H100 GPU
(80\,GB HBM3), median over 200 iterations after 50 warm-up iterations.

\begin{table}[t]
\caption{Forward-pass time (ms) on H100 CUDA, batch size 256.}
\label{tab:speed}
\begin{center}
\begin{tabular}{lrrrr}
\toprule
Method & dim=784 & dim=1024 & dim=2048 & dim=3072 \\
\midrule
Efficient-KAN (Cox-de Boor) & 0.722 & 0.721 & 0.732 & 0.919 \\
FastKAN (Gaussian RBF) & 0.230 & 0.229 & 0.241 & 0.256 \\
\textbf{InKAN (truncated power)} & \textbf{0.253} & \textbf{0.254} & \textbf{0.264} & \textbf{0.264} \\
\bottomrule
\end{tabular}
\end{center}
\end{table}

Under these configurations, InKAN has 2.8--3.5$\times$ lower
forward-pass latency than Efficient-KAN. FastKAN has the lowest latency
in all four configurations; InKAN's latency is approximately 3--11\%
higher.

\subsection{Classification}

Table~\ref{tab:accuracy} reports 10-epoch test accuracy on three
image classification datasets.

\begin{table}[t]
\caption{Test accuracy (\%) after 10 epochs, mean $\pm$ std over 10
seeds. B-spline variants use grid\_size=5, spline\_order=3. FastKAN
uses num\_grids=8 with Gaussian RBF basis.}
\label{tab:accuracy}
\begin{center}
\begin{tabular}{lrrr}
\toprule
Method & MNIST & FashionMNIST & CIFAR-10 \\
\midrule
MLP & \textbf{97.40}{\small$\pm$0.15} & 87.68{\small$\pm$0.43} & \textbf{50.23}{\small$\pm$0.65} \\
Efficient-KAN & 96.60{\small$\pm$0.24} & 87.55{\small$\pm$0.46} & 45.19{\small$\pm$0.63} \\
FastKAN & 97.23{\small$\pm$0.17} & \textbf{88.54}{\small$\pm$0.35} & 48.14{\small$\pm$0.58} \\
InKAN & 96.05{\small$\pm$0.27} & 86.92{\small$\pm$0.28} & 45.20{\small$\pm$0.49} \\
\bottomrule
\end{tabular}
\end{center}
\end{table}

MLP obtains the highest mean accuracy on MNIST and CIFAR-10. FastKAN
obtains the highest mean on FashionMNIST. B-spline KAN variants (InKAN,
Efficient-KAN) produce comparable results, with differences within
0.63\,pp across all datasets. These experiments compare the
implementations under the reported training configurations; they do
not isolate the causes of the observed differences.

\subsection{Regression}

Table~\ref{tab:regression} reports final test MSE on seven synthetic
functions after 3,000 epochs. Functions are: $\sin(x)$, $\cos(x)$,
$x^2$, $|x|$, $\exp(-x^2)$ (1D), $\sin(x_1 + x_2)$ (sin\_sum), and
$x_1 \cdot x_2$ (product, 2D). Training data are sampled from
$\mathcal{N}(0, 1)$.

\begin{table}[t]
\caption{Regression test MSE after 3,000 epochs, mean $\pm$ std over
10 seeds. FastKAN entries marked --- were not evaluated: the default
wrapper configuration rejected one-dimensional inputs with layer
normalization enabled. Bold indicates lowest mean among evaluated
methods.}
\label{tab:regression}
\begin{center}
\begin{tabular}{lrrrr}
\toprule
Function & InKAN & Efficient-KAN & FastKAN & MLP \\
\midrule
gaussian & \textbf{1.9e-5}{\tiny$\pm$1e-5} & 1.0e-4{\tiny$\pm$7e-5} & --- & 2.0e-3{\tiny$\pm$2e-3} \\
cos & \textbf{1.1e-3}{\tiny$\pm$4e-4} & 1.1e-3{\tiny$\pm$7e-4} & --- & 1.8e-3{\tiny$\pm$2e-3} \\
sin & 1.3e-3{\tiny$\pm$1e-3} & \textbf{7.9e-4}{\tiny$\pm$7e-4} & --- & 6.5e-3{\tiny$\pm$8e-3} \\
abs & 4.9e-3{\tiny$\pm$5e-3} & \textbf{3.8e-3}{\tiny$\pm$3e-3} & --- & 8.6e-3{\tiny$\pm$3e-3} \\
sin\_sum & 4.0e-3{\tiny$\pm$3e-3} & 4.4e-3{\tiny$\pm$4e-3} & 4.6e-2{\tiny$\pm$5e-3} & \textbf{1.0e-3}{\tiny$\pm$6e-4} \\
product & 3.6e-2{\tiny$\pm$3e-2} & 3.4e-2{\tiny$\pm$3e-2} & 1.4e-1{\tiny$\pm$4e-2} & \textbf{1.9e-2}{\tiny$\pm$1e-2} \\
$x^2$ & 9.8e-2{\tiny$\pm$1e-1} & 8.7e-2{\tiny$\pm$7e-2} & --- & \textbf{5.0e-2}{\tiny$\pm$5e-2} \\
\bottomrule
\end{tabular}
\end{center}
\end{table}

B-spline KAN variants achieve approximately $100\times$ lower MSE than
MLP on the Gaussian function in this configuration. MLP obtains the
lowest MSE on $x^2$, product, and sin\_sum. The default grid range
$[-1, 1]$ does not cover the $\mathcal{N}(0,1)$ input distribution;
in separate ablations with grid range $[-3, 3]$, KAN regression MSE
improved substantially on those functions.

\section{Related Work}
\label{sec:related}

\paragraph{KAN implementations.}
The original KAN~\citep{liu2024kan} evaluates B-spline bases via the
Cox-de Boor recursion. Efficient-KAN~\citep{blealtan2024} restructures
the computation into dense tensor operations.
FastKAN~\citep{li2024fastkan} replaces B-splines with Gaussian radial
basis functions. TruKAN~\citep{bayeh2026trukan} replaces B-splines with
raw truncated power functions, abandoning compact support and partition
of unity. InKAN preserves exact B-spline evaluation by using truncated
powers \emph{to compute} the basis, not as the basis itself.

\paragraph{Non-recursive and parallel spline evaluation.}
MatrixKAN~\citep{coffman2025matrixkan} parallelizes B-spline
computation using a matrix representation.
PowerMLP~\citep{qiu2025powermlp} connects B-splines to power-ReLU
networks. \citet{southworth2026multilevel} derived the same truncated
power stencil and used it for multilevel training.
ExSpliNet~\citep{fakhoury2022exsplinet} combines Kolmogorov-style
networks with multivariate B-spline representations. Lookup multivariate
KANs~\citep{pozdnyakov2025lookup} construct two-dimensional B-spline
surfaces in KAN layers. InKAN's contribution relative to these works is
the combination of bounded-coordinate stabilization
(Section~\ref{sec:bounded}) with compiler fusion and a production-ready
package.

\paragraph{Numerical stability of the truncated power form.}
De Boor~\citep{deboor2001practical} demonstrated that the truncated
power basis matrix is poorly conditioned. The bounded-coordinate
stabilization eliminates the problematic regime entirely, confining all
intermediate values to a bounded range (Table~\ref{tab:stability}).

\section{Discussion}
\label{sec:discussion}

\paragraph{Bounded-coordinate evaluation.}
Without the clamp, the truncated power form produces partition-of-unity
errors exceeding 0.88 at grid\_size=100 (Table~\ref{tab:stability}).
With the clamp, errors remain below $10^{-5}$ at all tested grid sizes
up to 200. Clamping prevents the growth of cancellation error associated
with large off-support coordinates. Small floating-point errors remain
within the support; their magnitude is bounded and quantified in
Table~\ref{tab:stability}.

\paragraph{Grid range alignment.}
When inputs leave the region covered by the spline grid, the spline
branch loses representational flexibility. Beyond the union of its basis
supports, that branch vanishes, while the residual branch
($w_{o,i}\sigma(x_i)$) can remain active. In ablation experiments,
extending the grid to cover a larger portion of the input distribution
improved regression MSE on the affected functions. Input normalization
and grid alignment are important practical considerations for KAN
performance.

\paragraph{Comparison with Gaussian RBF.}
FastKAN~\citep{li2024fastkan} introduced Gaussians to accelerate KANs.
Gaussians are non-negative and $C^\infty$, so the trade-off is not
smoothness. The properties lost are exact compact support and automatic
partition of unity. The truncated power form preserves both properties.
In the tested H100 configurations, FastKAN has slightly lower
forward-pass latency than InKAN (Table~\ref{tab:speed}).

\section{Limitations}
\label{sec:limitations}

The current implementation supports only cubic splines ($k{=}3$).
Setting other spline degrees raises an error. The closed form in
Eq.~\ref{eq:general_truncated} generalizes to arbitrary degree, but
only the cubic case has been implemented and tested.

The implemented fixed-coefficient formula assumes uniform knot spacing.
Non-uniform knot vectors require a different evaluator and are not
supported by this implementation.

Speed benchmarks report forward-pass latency only. Backward-pass timing,
peak memory measurements, and eager-versus-compiled fairness
comparisons are planned for a future revision.

\section{Conclusion}
\label{sec:conclusion}

InKAN evaluates uniform cubic B-spline basis functions via the truncated
power form with bounded-coordinate evaluation. The elementwise
expression is structured for \texttt{torch.compile} fusion. In the
tested H100 configurations, InKAN has 2.8--3.5$\times$ lower
forward-pass latency than the tested Cox-de Boor implementation.
Partition-of-unity errors remain below $10^{-5}$ for grid sizes up to
200. Classification and regression experiments show that B-spline KAN
variants using this evaluator produce comparable task performance to
other B-spline implementations under the reported configurations. The
package is open-source (\texttt{pip install inkan}, MIT license).

\bibliography{references}

@article{liu2024kan,
  title={KAN: Kolmogorov-Arnold Networks},
  author={Liu, Ziming and Wang, Yixuan and Vaidya, Sachin and Ruehle, Fabian and Halverson, James and Solja{\v{c}}i{\'c}, Marin and Hou, Thomas Y and Tegmark, Max},
  journal={arXiv preprint arXiv:2404.19756},
  year={2024}
}

@article{curry1947splines,
  title={On Spline Distributions and Their Limits: The {P}{\'o}lya Distribution Functions},
  author={Curry, Haskell B and Schoenberg, Isaac J},
  journal={Bulletin of the American Mathematical Society},
  volume={53},
  pages={1114},
  year={1947},
  note={Abstract 380t}
}

@article{schoenberg1946contributions,
  title={Contributions to the Problem of Approximation of Equidistant Data by Analytic Functions, {P}arts {A} and {B}},
  author={Schoenberg, Isaac J},
  journal={Quarterly of Applied Mathematics},
  volume={4},
  pages={45--99, 112--141},
  year={1946}
}

@misc{blealtan2024,
  title={An Efficient Implementation of {K}olmogorov-{A}rnold Network},
  author={Blealtan},
  year={2024},
  url={https://github.com/Blealtan/efficient-kan}
}

@article{li2024fastkan,
  title={{K}olmogorov-{A}rnold Networks are Radial Basis Function Networks},
  author={Li, Ziyao},
  journal={arXiv preprint arXiv:2405.06721},
  year={2024}
}

@article{cox1972numerical,
  title={The Numerical Evaluation of {B}-Splines},
  author={Cox, Maurice G},
  journal={Journal of the Institute of Mathematics and its Applications},
  volume={10},
  number={2},
  pages={134--149},
  year={1972}
}

@techreport{deboor1971subroutine,
  title={Subroutine Package for Calculating with {B}-Splines},
  author={de Boor, Carl},
  institution={Los Alamos Scientific Laboratory},
  number={LA-4728-MS},
  year={1971}
}

@book{deboor2001practical,
  title={A Practical Guide to Splines},
  author={de Boor, Carl},
  year={2001},
  edition={Revised},
  publisher={Springer-Verlag},
  address={New York}
}

@book{schumaker2007spline,
  title={Spline Functions: Basic Theory},
  author={Schumaker, Larry L},
  year={2007},
  edition={3rd},
  publisher={Cambridge University Press}
}

@article{southworth2026multilevel,
  title={Multilevel Training for {K}olmogorov-{A}rnold Networks},
  author={Southworth, Ben S and Actor, Jonas A and Harper, Graham and Cyr, Eric C},
  journal={arXiv preprint arXiv:2603.04827},
  year={2026}
}

@article{kolmogorov1957representation,
  title={On the Representation of Continuous Functions of Many Variables by Superposition of Continuous Functions of One Variable and Addition},
  author={Kolmogorov, Andrey N},
  journal={Doklady Akademii Nauk SSSR},
  volume={114},
  number={5},
  pages={953--956},
  year={1957}
}

@book{farin2002curves,
  title={Curves and Surfaces for {CAGD}: A Practical Guide},
  author={Farin, Gerald},
  year={2002},
  edition={5th},
  publisher={Morgan Kaufmann}
}

@article{coffman2025matrixkan,
  title={{MatrixKAN}: Parallelized {K}olmogorov-{A}rnold Network},
  author={Coffman, Cale and Chen, Lizhong},
  journal={arXiv preprint arXiv:2502.07176},
  year={2025}
}

@article{bayeh2026trukan,
  title={{TruKAN}: Towards More Efficient {K}olmogorov-{A}rnold Networks Using Truncated Power Functions},
  author={Bayeh, Ali and Sadaoui, Samira and Mouhoub, Malek},
  journal={arXiv preprint arXiv:2602.03879},
  year={2026}
}

@inproceedings{qiu2025powermlp,
  title={{PowerMLP}: An Efficient Version of {KAN}},
  author={Qiu, Ruichen and Miao, Yibo and Wang, Shiwen and Zhu, Yifan and Yu, Lijia and Gao, Xiao-Shan},
  booktitle={Proceedings of the AAAI Conference on Artificial Intelligence},
  volume={39},
  number={19},
  pages={20069--20076},
  year={2025},
  doi={10.1609/aaai.v39i19.34210}
}

@article{fakhoury2022exsplinet,
  title={{ExSpliNet}: An Interpretable and Expressive Spline-Based Neural Network},
  author={Fakhoury, Daniele and Fakhoury, Emanuele and Speleers, Hendrik},
  journal={Neural Networks},
  volume={152},
  pages={332--346},
  year={2022},
  doi={10.1016/j.neunet.2022.04.029}
}

@article{pozdnyakov2025lookup,
  title={Lookup Multivariate {K}olmogorov-{A}rnold Networks},
  author={Pozdnyakov, Sergey N and Schwaller, Philippe},
  journal={arXiv preprint arXiv:2509.07103},
  year={2025}
}
\bibliographystyle{iclr2027_conference}

\end{document}